\documentclass[final,5p,times,twocolumn]{elsarticle}

\usepackage{amsmath,amssymb}
\usepackage{graphicx}

\usepackage{multirow}
\usepackage{xcolor}
\usepackage{url}
\usepackage{hyperref}
\usepackage{booktabs}

\journal{Journal of Cultural Heritage}

\hypersetup{hidelinks}

\begin{document}

\begin{frontmatter}

\title{Compensating for Scarce Historical Images in Cross-Domain Cultural Heritage Retrieval Using Synthetic Aging}
\author[inst1,inst2]{Marcin Iwanowski\corref{cor1}}
\ead{marcin.iwanowski@pw.edu.pl}
\author[inst3]{Adam Mazgaj}
\ead{kontakt@clemens.pl}
\author[inst3]{Ferdynand Górski}
\ead{home@fgda.pl}
\author[inst4]{Sabina Szymoniak}
\ead{sabina.szymoniak@icis.pcz.pl}
\cortext[cor1]{Corresponding author.}
\address[inst1]{Inst.of Engineering and Technology, Faculty of Physics, Astronomy and Informatics, Nicolaus Copernicus University, ul.Grudziądzka 5, 87-100 Toruń, POLAND}
\address[inst2]{Institute of Control and Industrial Electronics, Warsaw University of Technology, ul.Koszykowa 75, 00-662 Warszawa, POLAND}
\address[inst3]{Clemens, ul.Pawlikowskiego 10/2, 31-127 Krakow, POLAND}
\address[inst4]{Department of Computer Science, Czestochowa University of Technology, ul.Dabrowskiego 69, 42-201 Czestochowa, POLAND}

\begin{abstract}
Cultural heritage collections often contain contemporary and historical visual records of the same physical object, but linking them is difficult because genuine historical images are scarce and differ from contemporary photographs. This study investigates whether synthetically aged contemporary images can replace or complement missing historical training observations in bidirectional instance-level retrieval.

A dataset of 1,077 identities, consisting of real old and contemporary images. Synthetic old-domain images were generated with a degradation-oriented aging pipeline. An EfficientNetV2-M model trained with batch-hard triplet loss was evaluated across three dataset partitions with pairwise disjoint test identities and three training seeds.

Complete synthetic replacement reduced the bidirectional mean R@1 from 92.15\% to 87.94\% (paired effect: $-4.21$ percentage points), while
increasing the synthetic pool from one to three variants produced no improvement. Under severe scarcity, synthetic completion was more
effective: at 10\% genuine old-domain coverage, mean R@1 improved by 12.30 and 13.99 points relative to exposure-matched and full-budget
real-only controls; at 30\%, the corresponding gains were 4.67 and 5.47 points. The benefit diminished as genuine coverage increased
and was negligible at 90\%. A prespecified 2-percentage-point acceptable-loss criterion was first satisfied at 70\% genuine coverage. Adding
synthetic observations at complete genuine coverage increased the mean R@1 to 94.05\%.

Synthetic aging is therefore more effective for completing missing cross-domain supervision than for replacing genuine historical data.
Broader identity coverage appears to be the primary benefit under severe scarcity, while a weaker augmentation effect may also be present.
\end{abstract}


\end{frontmatter}

\section{Introduction}
\label{sec:introduction}

Cultural heritage collections often contain multiple visual records of the same physical object created at different times and under different conditions. A painting, sculpture, icon, or museum artifact may be represented by a contemporary photograph, an archival image, a scanned catalog entry, or a printed reproduction. Such records are frequently dispersed across institutions, while differences in acquisition technology, digitization, color reproduction, resolution, framing, viewpoint, and physical degradation hinder direct visual correspondence. Automatically linking historical and contemporary records could support collection integration, catalog enrichment,
duplicate detection, provenance research, and the reconstruction of incomplete documentation.

This task can be formulated as a bidirectional cross-domain instance-level image retrieval. Given an image from one domain, the objective is to retrieve another representation of the same physical object from the opposite domain. Unlike generic similarity search, the target is not a semantically related object, but the same individual instance. Deep representations have become central to instance-level retrieval \cite{dubey2020cbirsurvey,radenovic2018revisiting}, while cultural heritage benchmarks demonstrate substantial appearance variation among images of the same artwork or museum object \cite{ypsilantis2022met,koniusz2018openmic}.

Cross-domain instance retrieval requires robustness to changes in viewpoint, scale, illumination, sharpness, color, background, framing, and image quality while preserving details that distinguish visually similar objects. Related generalization problems have been observed in cultural heritage image analysis \cite{Peaslee2025DomainGeneralizationPunchMarks} and are closely connected to domain adaptation and domain generalization \cite{ganin2016domain,zhou2023dgsurvey}. Deep metric learning provides a natural framework for this setting by constructing an embedding space in which images of the same object are close, and images of different objects are separated. Triplet-based objectives and hard-example mining have been widely used for instance recognition and retrieval \cite{hoffer2015triplet,schroff2015facenet,hermans2017triplet}. Their effectiveness, however, depends on representative positive pairs. Ideally, each training identity should therefore be observed in both the historical and contemporary domains.

This requirement is difficult to satisfy in cultural heritage collections. Contemporary photographs can often be acquired relatively easily, whereas historical records are scarce, unevenly distributed, poorly digitized, or difficult to locate. Some objects have extensive archival documentation, while others have only one historical image or none. Constructing a complete paired training set may consequently be infeasible even when contemporary imagery is abundant. The resulting imbalance is particularly problematic for metric-learning approaches, because identities without observations from both domains cannot provide genuine cross-domain positive pairs.

Synthetic aging offers a possible means of reducing this dependence. A contemporary photograph can be transformed into an identity-preserving old-domain representation using effects associated with historical photographs, catalog reproductions, scans, and degraded archival records. Related degradations have been studied in old photo restoration \cite{wan2020bringing,wan2020oldphoto}; here, they are deliberately introduced rather than removed to generate training observations. Synthetic aging can therefore be viewed not only as conventional data augmentation \cite{shorten2019survey}, but also as a mechanism
for completing an incompletely observed visual domain.

Its usefulness is not self-evident. A synthetically aged image retains the viewpoint, composition, and object geometry of its contemporary source and may therefore reproduce only part of the genuine historical-to-contemporary domain gap. Genuine historical records may contain structural and acquisition differences that cannot be reproduced by degradation-oriented transformations alone. Nevertheless, even imperfect synthetic observations may provide useful cross-domain supervision for identities lacking genuine historical counterparts.

The mechanism behind such an improvement is also ambiguous. Synthetic completion increases the number of identities represented in both domains, but it may simultaneously alter training exposure and introduce additional appearance variability. Consequently, an observed gain may arise from broader cross-domain identity coverage, from repeated exposure to a reduced genuine subset, or from a more general augmentation effect of synthetic old-domain observations. These mechanisms require separate controls because they lead to different practical conclusions about how synthetic data should be used when historical records are scarce.

Although synthetic imagery is widely used to enlarge training datasets, its role as a targeted substitute for missing identity--domain observations in cultural heritage retrieval remains underexplored. In particular, it is unclear how much genuine historical identity coverage is required before synthetic completion ceases to provide a meaningful benefit, whether multiple stochastic realizations of the same aging procedure improve synthetic-only training, and whether synthetic observations remain useful when genuine historical coverage is already complete. Addressing these questions require a controlled evaluation that varies genuine historical coverage while separating identity-coverage effects from training exposure and general augmentation.

\section{Research aim}
\label{sec:research_aim}

The aim of this study is to determine how synthetic aging can compensate for missing genuine historical observations in cross-domain cultural heritage retrieval and to identify the mechanism underlying any resulting improvement. We evaluate a degradation-based synthetic aging pipeline under complete replacement, partial synthetic completion, and full-coverage real-plus-synthetic training.

The study addresses four research questions:

\begin{description}
    \item[RQ1] To what extent can synthetically aged images replace genuine historical images in cross-domain retrieval training?

    \item[RQ2] Does increasing the number of independently generated synthetic variants improve retrieval when genuine historical observations are absent?

    \item[RQ3] How does retrieval performance change as genuine old-domain identity coverage varies and missing observations are replaced synthetically?

    \item[RQ4] Are the benefits of synthetic data explained primarily by broader cross-domain identity coverage, by repeated exposure to a reduced genuine subset, or by a more general augmentation effect when genuine coverage is already complete?
\end{description}

We hypothesize that synthetic aging cannot fully reproduce genuine historical variability, but can complement scarce genuine data by restoring missing cross-domain supervision, with its benefit diminishing as genuine historical coverage increases.

\section{Materials and methods}
\label{sec:method}

\subsection{Dataset and cross-domain retrieval scenario}
\label{sec:dataset}

The experiments use a cross-domain dataset of cultural heritage objects assembled from the Louvre Collections and the Lost Art Database \cite{LouvreCollections,LostArtDatabase}. Contemporary images define the \emph{new} domain, whereas original historical records and their corrected grayscale or sepia versions define the \emph{old} domain. The corrected versions preserve the original acquisition, viewpoint, and object representation and are therefore treated as genuine rather than synthetic observations.

Only identities represented in both domains are included. The final dataset contains 1077 object identities. Splitting is performed at the identity level, so all images belonging to a physical object are assigned to the same subset within a given dataset partition. Three partitions are evaluated. Each contains 808 training identities, 108 validation identities, and 161 test identities, corresponding approximately to a 75\%/10\%/15\% train--validation--test division.

The three partitions were constructed so that their test identity sets are pairwise disjoint. Consequently, the evaluation covers 483 distinct test identities across the three partitions. Training and validation identities are allowed to occur in more than one partition, provided that they do not overlap with the test set of the same partition. The partitions should therefore be regarded as three experimental realizations with distinct test identities rather than as statistically independent datasets.

Object-category information was used during partition construction to reduce differences in category composition between subsets. Test and validation identities were selected using category-stratified allocation, while preserving the required subset sizes and the pairwise disjointness of the test sets. Category-specific performance is not used as a primary outcome of the study.

Synthetic images are generated exclusively from contemporary images in the training subset of each partition. Validation and test sets contain only genuine old- and new-domain records. Reported retrieval performance therefore measures generalization across the genuine old--new domain gap and is never evaluated using synthetic queries or gallery images.

In the scarcity experiments, \emph{genuine old-domain identity coverage} denotes the proportion of training identities for which genuine old-domain observations are retained. In the real--synthetic completion conditions, identities without genuine old-domain observations receive synthetic old-domain counterparts, so that cross-domain training pairs remain available for the full training identity set. The construction of the corresponding real-only controls is described in Section~\ref{sec:experimental_design}.

\subsection{Synthetic aging of contemporary images}
\label{sec:synthetic_ageing}

Synthetic old-domain images are generated offline from contemporary training photographs using a degradation-oriented aging procedure. The objective is not to reconstruct the authentic historical appearance of a photograph, but to create identity-preserving old-domain observations that reproduce selected visual characteristics of historical photographs, printed reproductions, and digitized archival material.

For a contemporary image \(x_{\mathrm{new}}\) with identity \(y\), the synthetic representation is

\begin{equation}
\tilde{x}_{\mathrm{old}}
=
T_{\mathrm{age}}
\left(
x_{\mathrm{new}};\boldsymbol{\phi}
\right),
\qquad
y\left(\tilde{x}_{\mathrm{old}}\right)
=
y\left(x_{\mathrm{new}}\right),
\end{equation}

\noindent where \(\boldsymbol{\phi}\) contains independently sampled transformation parameters. The fixed transformation sequence is

\begin{equation}
\begin{aligned}
T_{\mathrm{age}} ={}&
T_{\mathrm{sepia}}
\circ
T_{\mathrm{leak}}
\circ
T_{\mathrm{paper}}
\circ
T_{\mathrm{persp}}
\circ
T_{\gamma}
\\
&\circ
T_{\mathrm{blur}}
\circ
T_{\mathrm{scratch}}
\circ
T_{\mathrm{vig}}
\circ
T_{\mathrm{noise}}
\circ
T_{\mathrm{frame}}.
\end{aligned}
\end{equation}

\begin{figure}[htbp]
    \centering
    \includegraphics[width=0.15\textwidth]{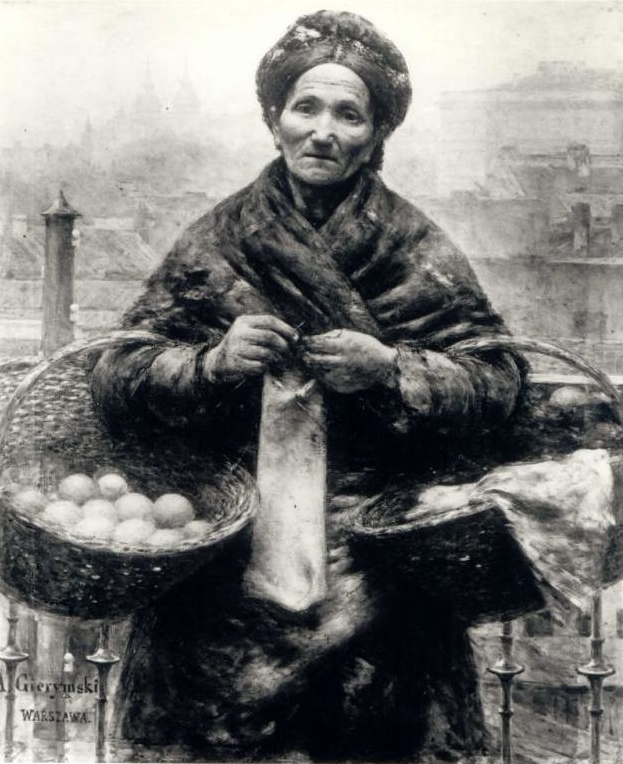}
    \hfill
    \includegraphics[width=0.155\textwidth]{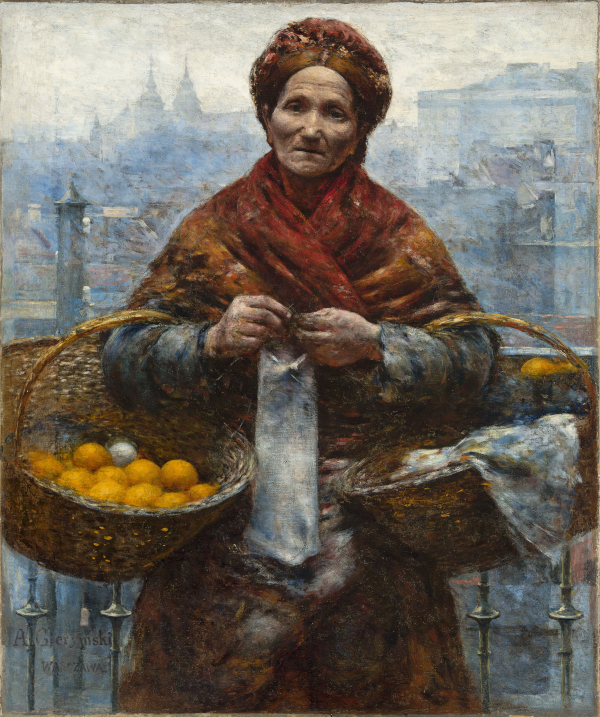}
    \hfill
    \includegraphics[width=0.162\textwidth]{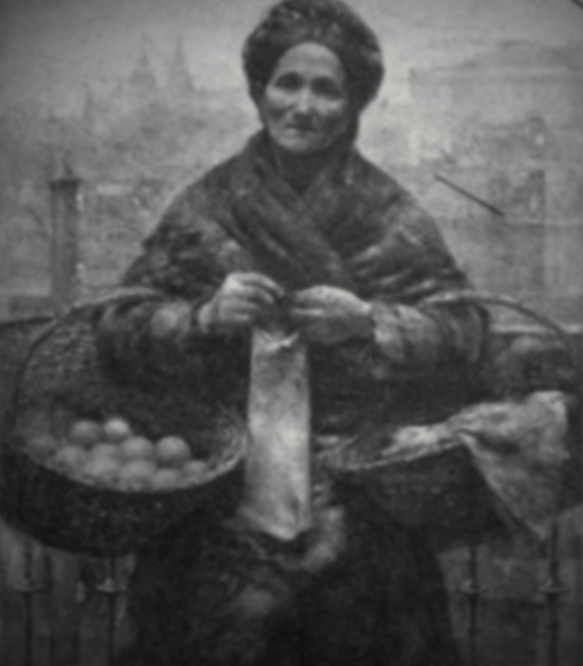}
    \caption{Aleksander Gierymski, Jewess with Oranges (Zydowka z pomarańczami), 1880–1881, National Museum in Warsaw. (a) Historical pre-1939 photograph; (b) contemporary digital reproduction; (c) synthetically aged version of (b), generated by the authors. The painting is in the public domain. Contemporary digital reproduction: National Museum in Warsaw/Wikimedia Commons, public domain.
    \label{fig:synthetic_aging_example}}
\end{figure}

The procedure introduces replicated borders, Gaussian and impulse noise, vignetting, scratches and dust artifacts, blur, tonal and perspective changes, paper texture, light artifacts, and sepia-to-grayscale conversion. It is inspired by degradations considered in old-photo restoration \cite{wan2020bringing,wan2020oldphoto}, but deliberately introduces rather than removes such effects. Figure~\ref{fig:synthetic_aging_example} compares a contemporary image, its genuine old-domain counterpart, and a synthetically aged variant.

Three synthetic variants, denoted OLDIFY1, OLDIFY2, and OLDIFY3, are generated using the same transformation pipeline and parameter ranges. They differ only in the independently sampled transformation parameters. Synthetic generation is performed separately for the training subset of each dataset partition. Consequently, when the same contemporary source image occurs in the training sets of more than one partition, it may receive a different synthetic realization in each partition.

Generation is deterministic for a fixed partition, source image, and OLDIFY variant, allowing the complete synthetic dataset to be reproduced. Within a given partition, the pre-generated synthetic images remain fixed across all training seeds and experimental conditions that use the corresponding OLDIFY variant. This preserves paired comparisons between conditions while allowing the three dataset partitions to represent different realizations of both data composition and synthetic degradation.

Synthetic images are generated only for training data and are never introduced into validation or test subsets. No online augmentation is applied during model training; input images are only resized and normalized using ImageNet statistics. Full transformation details, processing order, and parameter ranges are provided in Supplementary Methods~S1 and Table~S1.

\subsection{Retrieval model}
\label{sec:retrieval_model}

The retrieval model uses an EfficientNetV2-M backbone pretrained on ImageNet \cite{tan2021efficientnetv2,deng2009imagenet}. Its classification layer is replaced by a linear projection producing a 1280-dimensional, \(\ell_2\)-normalized embedding:

\begin{equation}
f_{\theta}(x) = \frac{Wg_{\theta}(x)+b} {\left\|Wg_{\theta}(x)+b\right\|_2},
\end{equation}

\noindent where \(g_{\theta}(x)\) denotes the backbone representation. The projection is initialized with \(W=I\) and \(b=0\), so the initial embedding corresponds to the normalized pretrained representation.

Training uses batch-hard triplet loss \cite{hoffer2015triplet,schroff2015facenet,hermans2017triplet}. For embeddings \(z_i=f_{\theta}(x_i)\), let \(d_{ij}=\lVert z_i-z_j\rVert_2\). The hardest positive and negative distances for anchor \(i\) are

\begin{equation}
d_i^{+} = \max_{\substack{j\neq i\\y_j=y_i}} d_{ij}, \qquad d_i^{-} = \min_{y_j\neq y_i} d_{ij}.
\end{equation}

The loss is

\begin{equation}
\mathcal{L}_{\mathrm{tri}} = \frac{1}{|\mathcal{B}|} \sum_{i\in\mathcal{B}} \max\left(0,d_i^{+}-d_i^{-}+m\right),
\end{equation}

\noindent where \(\mathcal{B}\) denotes the minibatch and \(m=0.3\).

Each batch contains eight identities and 16 images. For every selected identity, one old-domain and one new-domain image are sampled randomly from the corresponding image pools. Consequently, each identity contributes exactly one cross-domain positive pair per batch. When multiple old-domain variants are available, they enlarge the sampling pool but do not increase the number of images contributed by that identity in a single optimization step.

Input images are resized to \(384\times384\) pixels and normalized using ImageNet statistics. No online data augmentation is applied during training.

The final two backbone blocks and the projection layer are trainable, while all remaining backbone parameters are frozen. Backbone batch-normalization layers remain in evaluation mode and their parameters are not updated. AdamW is used with learning rates of \(10^{-5}\) for the unfrozen backbone and \(10^{-3}\) for the projection layer, weight decay \(10^{-4}\), and cosine annealing.

Training lasts for at most 30 epochs. Checkpoints are selected using bidirectional validation mean R@1, with an early-stopping patience of seven epochs and a minimum improvement of \(10^{-4}\). All conditions are trained with seeds 99, 100, and 101. Model architecture, optimization parameters, sampling structure, and stopping criteria are kept fixed across experimental conditions. The number of batches per epoch is determined from the available identity set according to the experimental protocol described in
Section~\ref{sec:experimental_design}.

\begin{table}[t]
\centering
\small
\caption{Main model and optimization parameters.}
\label{tab:model_training_parameters}
\begin{tabular}{ll}
\hline
Parameter & Value \\
\hline
Backbone & EfficientNetV2-M, ImageNet pretrained \\
Input resolution & \(384\times384\) \\
Embedding dimension & 1280 \\
Projection layer & Linear, identity initialization \\
Embedding normalization & \(\ell_2\) \\
Unfrozen backbone blocks & 2 final blocks \\
Backbone BatchNorm & Frozen \\
Loss function & Batch-hard triplet loss \\
Triplet margin & 0.3 \\
Identities per batch & 8 \\
Images per identity & 1 old and 1 new \\
Batch size & 16 \\
Maximum epochs & 30 \\
Optimizer & AdamW \\
Backbone learning rate & \(10^{-5}\) \\
Projection learning rate & \(10^{-3}\) \\
Weight decay & \(10^{-4}\) \\
Learning-rate schedule & Cosine annealing \\
Early-stopping criterion & Validation mean R@1 \\
Early-stopping patience & 7 epochs \\
No. of training seeds & 3 \\
Online augmentation & None \\
\hline
\end{tabular}
\end{table}

\subsection{Experimental design}
\label{sec:experimental_design}

Three experiments evaluate synthetic old-domain observations in complementary settings: complete replacement and full-coverage augmentation, synthetic completion under controlled scarcity of genuine historical observations, and real-only controls designed to separate the effect of broader identity coverage from increased training exposure.

All experiments use the same identity-based batch structure described in Section~\ref{sec:retrieval_model}. To avoid defining an epoch by an arbitrary fixed number of batches, the training budget is normalized with respect to the number of identities available to the sampler. Let \(N\) denote the number of training identities, \(P=8\) the number of identities sampled per batch, and \(B\) the number of batches per epoch. Assuming independent batch draws, the probability that a particular identity is sampled at least once during an epoch is

\begin{equation}
C(N,B)
=
1-
\left(
1-\frac{P}{N}
\right)^B .
\end{equation}

The full training schedule is defined by a target identity coverage of \(C_0=0.99\), giving

\begin{equation}
B_{\mathrm{full}}
=
\left\lceil
\frac{\log(1-C_0)}
{\log\left(1-P/N\right)}
\right\rceil .
\label{eq:full_batch_budget}
\end{equation}

For \(N=808\) training identities, this yields \(B_{\mathrm{full}}=463\) batches per epoch. This schedule is used whenever the full identity set is retained. The model architecture, optimizer, batch composition, early stopping, and all other training settings remain unchanged across conditions.

\subsubsection{Experiment 1: complete replacement and full-coverage augmentation}

Experiment~1 addresses RQ1 and RQ2. The \textsc{Real+Synth} condition additionally contributes to RQ4 by evaluating a general augmentation effect under complete genuine historical coverage. All four conditions use the full set of 808 training identities and \(B_{\mathrm{full}}=463\) batches per epoch. They differ only in the
source of the old-domain observation sampled for each identity.

The \textsc{Real} condition (\texttt{R100\_S0}) uses genuine old-domain images together with contemporary images. In \textsc{Synth-1} (\texttt{R0\_S100}), genuine old-domain images are completely replaced by OLDIFY1 variants. In \textsc{Synth-3}, the old-domain image is sampled from the combined pool of OLDIFY1, OLDIFY2, and OLDIFY3 variants. Because each selected identity still contributes only one old-domain image per batch, \textsc{Synth-1} and \textsc{Synth-3} differ in synthetic variability rather than in the number of optimization steps or images sampled per identity.

A fourth condition, \textsc{Real+Synth}, retains genuine old-domain observations for every training identity and additionally makes OLDIFY1 observations available. For each selected identity, the old-domain source is first chosen with equal probability from the genuine and synthetic pools, \(P(\mathrm{real})=P(\mathrm{synthetic})=0.5\), after which an image is sampled from the selected source. This source-balanced sampling prevents differences in pool cardinality from determining the relative frequency of genuine and synthetic observations. Comparison with \textsc{Real} tests whether synthetic aging provides useful augmentation even when no historical observations are missing.

\subsubsection{Experiment 2: controlled scarcity and synthetic completion}

Experiment~2 addresses RQ3 by varying the proportion of training identities for which genuine old-domain observations are retained. Nine intermediate genuine-coverage levels are evaluated,

\[ r \in \{10,20,30,40,50,60,70,80,90\}\%.\]

The corresponding mixed condition is denoted \(\texttt{R}r\texttt{\_S}(100-r)\). For each partition, the number of genuine identities is

\begin{equation}
N_{\mathrm{real}} = \operatorname{round} \left( \frac{r}{100}N \right),
\end{equation}

\noindent and all remaining training identities receive OLDIFY1 old-domain observations. Each identity is assigned exclusively to one old-domain source: all available genuine old-domain observations are used for identities in the genuine subset, whereas identities outside this subset use their OLDIFY1 counterparts. Contemporary images remain available for all 808 identities. Thus, every mixed condition preserves the full cross-domain training identity set.

The genuine subsets are nested,

\[R10 \subset R20 \subset \cdots \subset R90 \subset R100, \]

\noindent and are constructed separately for each dataset partition. The ordering is category-stratified so that successive prefixes approximately preserve the category composition of the full training set. The resulting source assignment is fixed across all three training seeds within a partition. Consequently, differences between model seeds do not alter which identities receive genuine or synthetic old-domain observations.

All mixed conditions retain \(N=808\) training identities and use \(B_{\mathrm{full}}=463\) batches per epoch. The endpoints of the coverage curve are reused from Experiment~1: \texttt{R0\_S100} corresponds to \textsc{Synth-1}, whereas \texttt{R100\_S0} corresponds to \textsc{Real}. The complete coverage series therefore spans genuine old-domain identity coverage from 0\% to 100\% in 10-percentage-point increments.

\subsubsection{Experiment 3: real-only controls for identity coverage and training exposure}

Experiment~3 addresses RQ4 using real-only controls at
 
\[ r \in \{10,30,50,70,90\}\%. \]

At each coverage level, the corresponding mixed condition and both real-only controls use exactly the same genuine identity subset. Unlike the mixed condition, however, the real-only controls remove all remaining identities entirely from training, including their contemporary images. Hence, the A and B variants contain only \(N_{\mathrm{real}}\) identities represented by genuine old- and new-domain observations.

The A variants (\(\texttt{R}r\texttt{\_A}\)) use an exposure-matched schedule obtained by scaling the full-set budget according to the number of retained identities:

\begin{equation}
B_A
=
\operatorname{round}
\left(
B_{\mathrm{full}}
\frac{N_{\mathrm{real}}}{N}
\right).
\label{eq:a_batch_budget}
\end{equation}

This preserves, to rounding accuracy, the expected number of selections of each retained identity per epoch relative to the full-set schedule. For 10\%, 30\%, 50\%, 70\%, and 90\% genuine coverage, the resulting schedules contain 46, 139, 232, 324, and 417 batches per epoch, respectively. 

The B variants (\(\texttt{R}r\texttt{\_B}\)) use the same reduced genuine identity sets but retain the full \(B_{\mathrm{full}}=463\)-batch schedule. They therefore match the corresponding mixed conditions in the number of optimization steps and image selections per epoch while repeatedly sampling a smaller set of genuine identities.

Three paired comparisons are used to distinguish the evaluated effects. Comparing a mixed condition with its A control measures the benefit of synthetic completion relative to real-only training with matched expected per-identity exposure. Comparing B with A estimates the effect of increasing repeated exposure to the same reduced genuine subset. Finally, comparing the mixed condition with B holds the number of batches and per-epoch image selections constant and tests whether retaining broader cross-domain identity coverage provides an advantage over repeated sampling of the reduced genuine subset.

\begin{table}[t]
\centering
\scriptsize
\caption{Summary of the experimental training conditions.
\(N=808\) denotes the full training identity set,
\(N_{\mathrm{real}}=\operatorname{round}(rN/100)\), and
\(B_{\mathrm{full}}=463\).}
\label{tab:experimental_conditions}
\begin{tabular}{lllll}
\hline
Exp. &
Condition &
Genuine coverage &
Training IDs &
Batches/epoch \\

\hline
1 &
\textsc{Real} (\texttt{R100\_S0}) &
100\% &
\(N\) &
\(B_{\mathrm{full}}\) \\

1 &
\textsc{Synth-1} (\texttt{R0\_S100}) &
0\% &
\(N\) &
\(B_{\mathrm{full}}\) \\

1 &
\textsc{Synth-3} &
0\% &
\(N\) &
\(B_{\mathrm{full}}\) \\

1 &
\textsc{Real+Synth} &
100\% genuine + synthetic &
\(N\) &
\(B_{\mathrm{full}}\) \\

2 &
\(\texttt{R}r\texttt{\_S}(100-r)\) &
\(r=10,20,\ldots,90\%\) &
\(N\) &
\(B_{\mathrm{full}}\) \\

3 &
\(\texttt{R}r\texttt{\_A}\) &
\(r=10,30,50,70,90\%\) &
\(N_{\mathrm{real}}\) &
\(B_A\) \\

3 &
\(\texttt{R}r\texttt{\_B}\) &
\(r=10,30,50,70,90\%\) &
\(N_{\mathrm{real}}\) &
\(B_{\mathrm{full}}\) \\
\hline
\end{tabular}
\end{table}

Across the three dataset partitions and three training seeds, the protocol required 207 separate model training runs: 36 for Experiment~1 and 171 for Experiments~2--3. The \texttt{R0\_S100} and \texttt{R100\_S0} endpoints of the scarcity series reuse the corresponding Experiment~1 models and were not retrained.

\subsection{Retrieval evaluation}
\label{sec:retrieval_evaluation}

All models are evaluated using the same bidirectional instance-level retrieval protocol \cite{radenovic2018revisiting}. For each query, similarity is computed against every test image from the opposite domain. Because the embeddings are \(\ell_2\)-normalized, similarity is defined as

\begin{equation}
s(q,g) = f_{\theta}(q)^{\top}f_{\theta}(g),
\end{equation}

\noindent which is equivalent to cosine similarity.

In old-to-new retrieval, genuine old-domain images are used as queries and contemporary images form the gallery. In new-to-old retrieval, contemporary images are used as queries and genuine old-domain images form the gallery. Only genuine observations are used for evaluation; synthetic images are never included as queries or gallery items.

Because an identity may be represented by multiple gallery images, retrieval at rank \(k\) is considered successful when at least one image of the correct identity occurs among the first \(k\) retrieved items:

\begin{equation}
\mathrm{R@}k = \frac{1}{|\mathcal{Q}|} \sum_{q\in\mathcal{Q}} \mathbf{1} \left[ \exists g\in\operatorname{Top}_{k}(q): y_g=y_q \right],
\end{equation}

\noindent where \(\mathcal{Q}\) denotes the query set. Thus, \(\mathrm{R@}k\) is query-weighted: every query contributes equally within a retrieval direction, irrespective of how many images are available for its identity.

Recall is reported for \(k\in\{1,5,10\}\). The primary evaluation metric is bidirectional mean R@1,

\begin{equation}
\mathrm{mean\ R@1} = \frac{\mathrm{R@1}_{\mathrm{old}\rightarrow\mathrm{new}} + \mathrm{R@1}_{\mathrm{new}\rightarrow\mathrm{old}}}{2}.
\end{equation}

This metric assigns equal weight to the two retrieval directions and is also used for checkpoint selection on the validation subset. R@5 and R@10 provide complementary measures of whether the correct identity remains within a short candidate list suitable for expert
inspection.

Within each dataset partition, all compared models are evaluated using exactly the same genuine queries and galleries. This permits paired comparisons between experimental conditions at the query level while ensuring that differences in retrieval performance cannot result from changes in the evaluation set.

\subsection{Statistical analysis}
\label{sec:statistical_analysis}

Each experimental condition is evaluated in three dataset partitions and with three training seeds, yielding nine split--seed results. For descriptive reporting, retrieval performance is summarized as the mean and sample standard deviation across these nine runs. Because the three seeds within a partition share the same test identities, and because training and validation identities may overlap between partitions, the nine results are not treated as independent observations for confidence-interval estimation.

Comparisons between experimental conditions are paired within partition and training seed. Since all models compared within a partition are evaluated using identical queries and galleries, paired effects can be computed directly at the query level. For a metric \(M\), the effect of variant \(B\) relative to reference \(A\) is defined as

\begin{equation}
\Delta M = M_B-M_A.
\end{equation}

Confidence intervals for paired differences in the primary metric, bidirectional mean R@1, are estimated using 10,000 paired bootstrap resamples at the test-identity level. Within each dataset partition, test identities are sampled with replacement. For identities represented by more than one query in a retrieval direction, all queries belonging to the selected
identity are retained together. Training seeds are independently resampled with replacement within the same partition.

For every bootstrap replicate, query-weighted R@1 differences are calculated separately for old-to-new and new-to-old retrieval, averaged across the two directions, and then averaged across the three dataset partitions with equal partition weight. The partitions themselves are not resampled. The 95\% confidence interval is given by the 2.5th and 97.5th percentiles
of the bootstrap distribution.

This procedure preserves the query-weighted definition of R@1 from Section~3.5 while using object identity as the resampling unit. The great majority of test identities are represented by one old-domain and one new-domain image; for the small number of identities represented by multiple images, retaining all associated queries together prevents pseudo-replication. Resampling training seeds additionally accounts for variability due to model training.

For Experiment~1, the principal paired comparisons are

\[ \textsc{Synth-1}-\textsc{Real},\qquad \textsc{Synth-3}-\textsc{Real},\qquad \textsc{Synth-3}-\textsc{Synth-1}, \]

\noindent together with

\[ \textsc{Real+Synth}-\textsc{Real}, \]

\noindent which evaluates whether synthetic old-domain observations provide an augmentation benefit when genuine historical coverage is already complete. 

For the scarcity controls in Experiment~3, three paired effects are evaluated at genuine old-domain coverage levels \(r\in\{10,30,50,70,90\}\%\):

\begin{align}
\Delta_{\mathrm{completion},A}
&=
M_{\mathrm{R}r\mathrm{\_S}(100-r)}
-
M_{\mathrm{R}r\mathrm{\_A}},
\\
\Delta_{\mathrm{exposure}}
&=
M_{\mathrm{R}r\mathrm{\_B}}
-
M_{\mathrm{R}r\mathrm{\_A}},
\\
\Delta_{\mathrm{completion},B}
&=
M_{\mathrm{R}r\mathrm{\_S}(100-r)}
-
M_{\mathrm{R}r\mathrm{\_B}}.
\end{align}

The first comparison measures the benefit of synthetic completion relative to real-only training with matched expected per-identity exposure. The second measures the effect of increasing repeated exposure to the same reduced genuine identity subset. The third compares mixed and real-only training under the same full per-epoch optimization budget.

Mixed completion conditions are additionally compared with the complete genuine-data reference \texttt{R100\_S0}. Before inspection of the results from the revised experimental protocol, an acceptable-loss margin of

\begin{equation}
\delta = 0.02
\end{equation}

\noindent was specified, corresponding to a maximum reduction of two percentage points in bidirectional mean R@1. For a mixed condition,

\begin{equation}
\Delta_r
=
\mathrm{mean\ R@1}_{\mathrm{R}r\mathrm{\_S}(100-r)}
-
\mathrm{mean\ R@1}_{\mathrm{R100\_S0}},
\end{equation}

\noindent the descriptive acceptable-loss criterion is satisfied when

\begin{equation}
\Delta_{r,\mathrm{low}}>-\delta,
\end{equation}

\noindent where \(\Delta_{r,\mathrm{low}}\) is the lower bound of the 95\% bootstrap confidence interval. This analysis is used to identify the lowest evaluated genuine-coverage level whose uncertainty interval remains within the prespecified two-percentage-point margin. It is not interpreted as a formal non-inferiority test.

The three dataset partitions have pairwise disjoint test identities but originate from the same assembled collection and may share training or validation identities. Confidence intervals therefore quantify uncertainty associated with the evaluated test identities and training seeds rather than population-level uncertainty across independent datasets. Partition-level results are additionally examined to assess whether the principal effects are consistent across different test-set compositions.

\section{Results}
\label{sec:results}

\subsection{Complete replacement and full-coverage augmentation}
\label{sec:results_replacement}

Table~\ref{tab:results_replacement} summarizes Experiment~1, including complete synthetic replacement, increased synthetic variability, and synthetic augmentation under complete genuine old-domain coverage.

\begin{table}[t]
\centering
\scriptsize
\caption{Retrieval performance in Experiment~1, reported as mean $\pm$ sample standard deviation over three dataset partitions and three training seeds.}
\label{tab:results_replacement}
\begin{tabular}{lccc}
\hline
Variant &
Old$\rightarrow$New R@1 &
New$\rightarrow$Old R@1 &
Mean R@1 \\
\hline
\textsc{Real}
& $92.21 \pm 3.44$\%
& $92.10 \pm 3.73$\%
& $92.15 \pm 3.52$\% \\

\textsc{Synth-1}
& $88.19 \pm 2.50$\%
& $87.69 \pm 3.12$\%
& $87.94 \pm 2.68$\% \\

\textsc{Synth-3}
& $88.06 \pm 3.45$\%
& $87.76 \pm 3.29$\%
& $87.91 \pm 3.29$\% \\

\textsc{Real+Synth}
& $94.19 \pm 2.52$\%
& $93.91 \pm 3.08$\%
& $94.05 \pm 2.70$\% \\
\hline
\end{tabular}
\end{table}

Complete replacement of genuine old-domain observations with OLDIFY1 reduced bidirectional mean R@1 from 92.15\% for \textsc{Real} to 87.94\% for \textsc{Synth-1}. The paired effect was $-4.21$ percentage points (pp), with a 95\% bootstrap confidence interval (CI) of $[-6.19,-2.23]$~pp. The reduction was observed in both retrieval directions, showing that synthetic aging
did not reproduce the full retrieval value of genuine historical observations.

Increasing the synthetic sampling pool from one to three independently generated OLDIFY variants produced essentially no change in rank-1 retrieval. \textsc{Synth-3} achieved 87.91\% mean R@1, and its paired difference relative to \textsc{Synth-1} was $-0.03$~pp (95\% CI: $[-2.39,+2.43]$~pp). Relative to \textsc{Real}, the difference for \textsc{Synth-3} was
$-4.24$~pp (95\% CI: $[-6.93,-1.71]$~pp). Increasing the number of stochastic realizations of the same degradation pipeline therefore did not reduce the gap between synthetic and genuine old-domain training observations.

Synthetic-only training nevertheless retained strong retrieval performance beyond rank~1. \textsc{Synth-1} achieved a bidirectional mean R@10 of 98.43\%, compared with 99.23\% for \textsc{Real}. Thus, complete synthetic replacement usually placed the correct identity within a short candidate list even though it was less effective at ranking the correct identity first.

The \textsc{Real+Synth} condition produced the highest mean R@1, 94.05\%, compared with 92.15\% for \textsc{Real}. The paired difference was $+1.90$~pp, with a 95\% CI of $[-0.10,+4.09]$~pp. The interval slightly crossed zero, so the result does not provide clear evidence of an augmentation benefit under complete genuine old-domain coverage. Nevertheless, the
positive point estimate in both retrieval directions suggests that synthetic old-domain observations may provide additional useful appearance variability even when genuine historical observations are available for every training identity.

\subsection{Effect of genuine old-domain coverage and synthetic completion}
\label{sec:results_scarcity}

\begin{figure*}[t]
    \centering
    \begin{minipage}[t]{0.49\textwidth}
        \centering
        \textbf{(a)}\\[-0.5ex]
        \includegraphics[width=\linewidth]{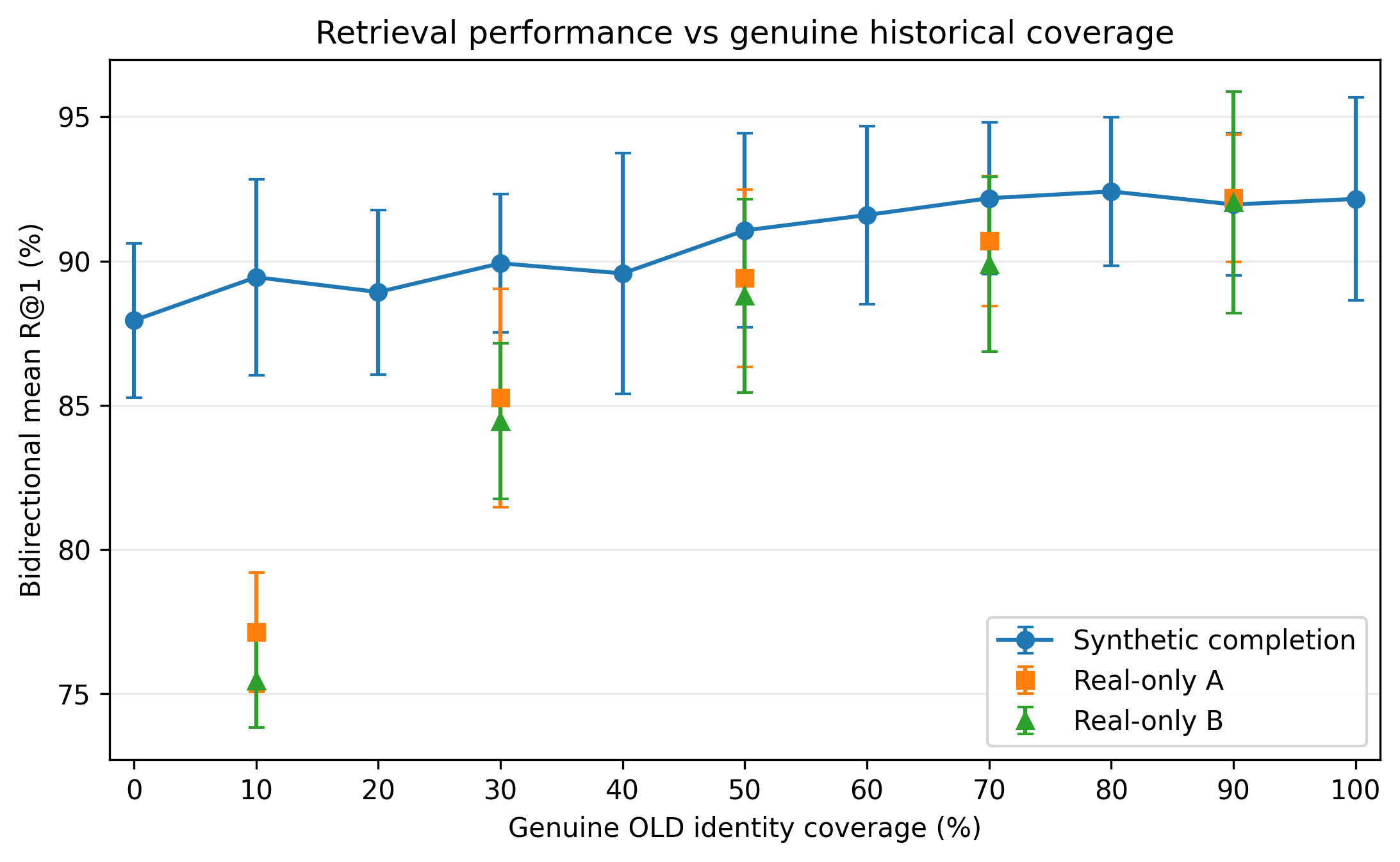}
    \end{minipage}
    \hfill
    \begin{minipage}[t]{0.49\textwidth}
        \centering
        \textbf{(b)}\\[-0.5ex]
        \includegraphics[width=\linewidth]{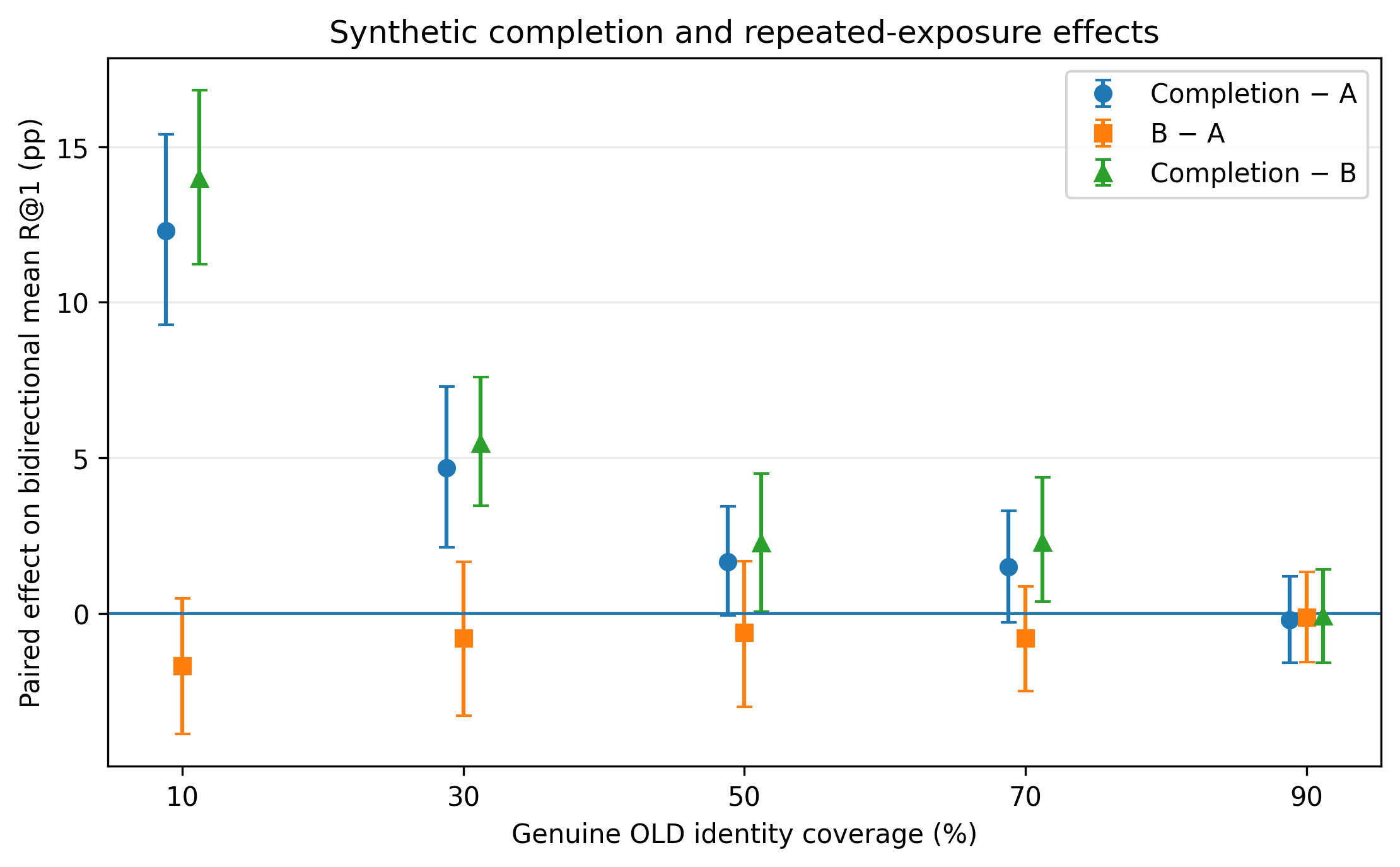}
    \end{minipage}
    \caption{Effect of genuine old-domain identity coverage on bidirectional mean R@1. (a) Retrieval performance for the real--synthetic completion series and the real-only A and B controls. Mixed conditions retain all training identities and synthetically complete those without genuine old-domain observations. A controls use the reduced genuine identity set with matched expected per-identity exposure, whereas B controls use the same reduced identity set with the full per-epoch optimization budget. Error bars in panel~(a) represent sample standard deviations across three dataset partitions and three training seeds. (b) Paired effects of synthetic completion and increased repeated exposure. Error bars in panel (b) represent 95\% confidence intervals obtained using the paired identity-level bootstrap described in Section 3.6.}
    \label{fig:scarcity_results}
\end{figure*}

Figure~\ref{fig:scarcity_results} summarizes retrieval performance under progressively increasing genuine old-domain identity coverage. Panel~(a) shows the complete synthetic-completion series from 0\% to 100\% genuine coverage in 10-percentage-point increments, together with the real-only A and B controls evaluated at 10\%, 30\%, 50\%, 70\%, and 90\%. Panel~(b) shows the corresponding paired effects; these comparisons are analyzed in detail in Section~\ref{sec:results_paired}. Bidirectional mean R@1 values for all scarcity and real-only control conditions are provided in Supplementary Table~\ref{tab:supp_scarcity_results}.

Across the mixed real--synthetic series, retrieval performance generally increased as genuine old-domain coverage increased, although the relationship was not strictly monotonic. Complete synthetic replacement (\texttt{R0\_S100}) achieved 87.94\% mean R@1. Introducing 10\% genuine coverage increased mean R@1 to 89.44\%, while the 30\% and 50\% conditions reached 89.92\% and 91.06\%, respectively. Performance approached the complete-genuine-data reference at higher coverage levels: \texttt{R70\_S30} achieved 92.18\%, \texttt{R80\_S20} 92.41\%, and \texttt{R90\_S10} 91.96\%, compared with 92.15\% for \texttt{R100\_S0}. Thus, the mixed conditions from 70\% to 90\% genuine coverage remained within approximately 0.3 percentage points of the complete-genuine-data mean.

The advantage of retaining the full training identity set was most pronounced when genuine historical observations were highly scarce. At 10\% genuine coverage, \texttt{R10\_S90} achieved 89.44\% mean R@1, whereas the corresponding real-only A and B controls achieved 77.14\% and 75.45\%, respectively. At 30\% coverage, the mixed condition achieved 89.92\%, compared with 85.26\% for \texttt{R30\_A} and 84.45\% for \texttt{R30\_B}. The separation became smaller as genuine coverage increased. At 50\%, the respective values were 91.06\%, 89.41\%, and 88.79\%, and at 70\% they were 92.18\%, 90.69\%, and 89.89\%. By 90\% genuine coverage, the three conditions converged, with mean R@1 values of 91.96\% for the mixed condition, 92.18\% for A, and 92.04\% for B.
 
The overall pattern therefore indicates a diminishing performance advantage of synthetic completion as genuine old-domain coverage increases. The largest separation from the real-only controls occurs under severe historical-data scarcity, whereas little difference remains when genuine observations are available for nearly all training identities. The magnitude and uncertainty of these paired effects are examined next.

\subsection{Paired effects and acceptable-loss analysis}
\label{sec:results_paired}

Figure~\ref{fig:scarcity_results}(b) summarizes the paired effects of synthetic completion and increased repeated exposure. Numerical effect estimates and 95\% bootstrap confidence intervals are reported in Table~\ref{tab:paired_differences}.

\begin{table}[t]
\centering
\scriptsize
\caption{Paired differences in bidirectional mean R@1 for the scarcity controls. Positive values favor the first variant. Confidence intervals were estimated using the paired identity-level bootstrap with seed resampling described in
Section~3.6.}
\label{tab:paired_differences}
\begin{tabular}{llcc}
\hline
Comparison type &
Coverage &
Mean difference &
95\% CI \\
\hline

Completion vs.\ A
& 10\%
& $+12.30$ pp
& $[+9.27,+15.40]$ pp \\

& 30\%
& $+4.67$ pp
& $[+2.11,+7.28]$ pp \\

& 50\%
& $+1.65$ pp
& $[-0.06,+3.45]$ pp \\

& 70\%
& $+1.48$ pp
& $[-0.29,+3.29]$ pp \\

& 90\%
& $-0.22$ pp
& $[-1.59,+1.19]$ pp \\
\hline

Additional exposure
& 10\%
& $-1.68$ pp
& $[-3.88,+0.47]$ pp \\

& 30\%
& $-0.81$ pp
& $[-3.29,+1.66]$ pp \\

& 50\%
& $-0.62$ pp
& $[-3.02,+1.67]$ pp \\

& 70\%
& $-0.80$ pp
& $[-2.51,+0.87]$ pp \\

& 90\%
& $-0.14$ pp
& $[-1.57,+1.32]$ pp \\
\hline

Completion vs.\ B
& 10\%
& $+13.99$ pp
& $[+11.23,+16.82]$ pp \\

& 30\%
& $+5.47$ pp
& $[+3.45,+7.60]$ pp \\

& 50\%
& $+2.27$ pp
& $[+0.06,+4.49]$ pp \\

& 70\%
& $+2.29$ pp
& $[+0.37,+4.36]$ pp \\

& 90\%
& $-0.08$ pp
& $[-1.60,+1.41]$ pp \\
\hline
\end{tabular}
\end{table}

Synthetic completion provided its largest advantage under severe historical-data scarcity. At 10\% genuine old-domain coverage, the mixed condition exceeded the exposure-matched A control by 12.30~pp (95\% CI: $[+9.27,+15.40]$~pp) and the full-budget B control by 13.99~pp (95\% CI: $[+11.23,+16.82]$~pp). At 30\% coverage, the corresponding gains remained substantial, at
4.67~pp and 5.47~pp, and both confidence intervals remained above zero.

The completion advantage decreased as genuine coverage increased. At 50\%, the mixed condition exceeded A by 1.65~pp, with the confidence interval narrowly including zero, whereas its 2.27~pp advantage over B had a lower confidence bound just above zero. A similar pattern was observed at 70\% coverage: the mixed--A effect was $+1.48$~pp with a confidence interval including
zero, while the mixed--B effect was $+2.29$~pp (95\% CI: $[+0.37,+4.36]$~pp). At 90\%, the differences between the mixed and real-only variants were close to zero.

In contrast, increasing the training budget for the reduced real-only identity sets did not reproduce the completion effect. All B--A point estimates were small and negative, ranging from $-1.68$ to $-0.14$~pp, and every confidence interval included zero. Thus, the large gains observed at low genuine coverage cannot be explained by additional optimization steps or repeated exposure to the same reduced genuine identity subset.

The mixed completion series was also compared with the complete genuine-data reference \texttt{R100\_S0} using the prespecified two-percentage-point acceptable-loss margin. At 50\% genuine coverage, the paired effect relative to \texttt{R100\_S0} was $-1.09$~pp, but the lower confidence bound reached $-2.68$~pp. At 60\%, the effect was $-0.56$~pp with a 95\% CI of
$[-2.19,+1.15]$~pp. At 70\%, the effect was $+0.03$~pp and the 95\% CI was $[-1.78,+1.88]$~pp. Consequently, 70\% genuine old-domain identity coverage was the lowest evaluated level whose 95\% bootstrap confidence interval remained within the prespecified 2-pp acceptable-loss margin relative to complete genuine coverage. The 80\% and 90\% conditions also remained within
this margin.

The complete acceptable-loss curve is reported in Supplementary Figure~S2. This analysis is descriptive and is not interpreted as a formal non-inferiority test.

\subsection{Variability across dataset partitions}
\label{sec:results_partitions}

Absolute retrieval performance varied across the three dataset partitions, reflecting differences in test-set composition. Nevertheless, the principal scarcity-related pattern was reproduced across the pairwise disjoint test sets.

The strongest consistency was observed at low genuine historical coverage. For the mixed condition relative to the exposure-matched A control, the split-level gains at 10\% coverage were $+10.09$, $+14.63$, and $+12.19$~pp for Splits~1--3, respectively. At 30\% coverage, the corresponding gains were $+6.29$, $+4.36$, and $+3.35$~pp. Thus, synthetic completion
provided a clear advantage in every partition when genuine old-domain observations were highly scarce.

The effect became smaller at higher coverage levels. At 50\%, the mixed--A differences were $+0.87$, $+1.30$, and $+2.79$~pp, and at 70\% they were $+0.76$, $+2.06$, and $+1.63$~pp. By 90\% coverage, the completion effect was small and its direction was no longer consistent across partitions.

These results show that dataset composition affects both absolute retrieval difficulty and the magnitude of the synthetic-completion effect. However, the decreasing benefit of synthetic completion as genuine historical coverage increases is reproduced across all three test partitions. Detailed partition-level results and the corresponding split-specific completion curves are provided in Supplementary Section~S2.

\section{Discussion}
\label{sec:discussion}

The results distinguish three roles of synthetic old-domain observations: replacement of genuine historical data, completion of missing cross-domain identity observations, and augmentation when genuine historical coverage is already complete. These roles produce different effects. Complete synthetic replacement remained inferior to genuine historical training data, whereas synthetic completion provided large gains when genuine old-domain observations were scarce. Increasing repeated exposure to the same reduced genuine identity subsets did not reproduce these gains. At complete genuine coverage, however, adding synthetic observations produced a smaller positive effect whose confidence interval narrowly included zero. Taken together, the results indicate that broader cross-domain training identity coverage is a primary contributor to the benefit of synthetic data under severe scarcity, while a weaker general augmentation effect may also be present.

\subsection{Complete replacement and synthetic diversity}
\label{sec:discussion_replacement}

Experiment~1 addresses RQ1 by showing that synthetically aged contemporary images are not equivalent to genuine historical observations. Complete replacement with OLDIFY1 reduced bidirectional mean R@1 by 4.21 percentage points, from 92.15\% to 87.94\%, with a 95\% bootstrap confidence interval of $[-6.19,-2.23]$~pp. The corresponding reduction was observed in
both retrieval directions.

This gap is consistent with the limitations of a degradation-oriented transformation. OLDIFY modifies noise, blur, scratches, tone, perspective, paper appearance, light artifacts, and other low-level characteristics, but the resulting image is still derived from the contemporary source. It therefore largely preserves its viewpoint, composition, object geometry, and scene structure. Genuine historical records may additionally differ in acquisition viewpoint, crop, framing, background, reproduction process, physical condition, and digitization. Such differences cannot be reproduced fully by applying appearance degradations to a contemporary photograph.

Synthetic-only training nevertheless remained informative. Although its rank-1 performance was lower than that obtained with genuine historical observations, \textsc{Synth-1} achieved a bidirectional mean R@10 of 98.43\%, compared with 99.23\% for \textsc{Real}. Synthetic aging therefore often placed the correct object within a short candidate list even when it did not provide sufficient cross-domain variability to rank the correct identity first. This distinction is relevant for cultural heritage workflows in which retrieval is used to generate candidates for subsequent expert verification rather than to make fully automatic identity decisions.

Experiment~1 also addresses RQ2. Increasing the old-domain sampling pool from one to three independently generated OLDIFY variants changed mean R@1 by only $-0.03$~pp, with a 95\% confidence interval of $[-2.39,+2.43]$~pp. Thus, additional stochastic realizations of the same degradation pipeline did not reduce the gap to genuine historical data. Increasing synthetic multiplicity alone appears less important than introducing forms of variability not present in the current transformation family.

Future improvements to synthetic replacement may therefore require methods that alter more structural properties of the observation, including viewpoint, framing, background, acquisition style, or visible object condition, rather than generating additional
realizations of similar low-level degradations.

\subsection{Synthetic completion under historical-data scarcity}
\label{sec:discussion_completion}

Experiments~2 and~3 address RQ3 and the main identity-coverage and training-exposure components of RQ4. The mixed completion series showed an overall increase in retrieval performance as genuine old-domain coverage increased, although the relationship was not strictly monotonic. The benefit of synthetic completion was largest when genuine historical observations were scarce and diminished substantially as genuine coverage approached completeness.

The strongest effects occurred at 10\% genuine coverage. The mixed condition exceeded the exposure-matched A control by 12.30~pp (95\% CI: $[+9.27,+15.40]$~pp) and the full-budget B control by 13.99~pp (95\% CI: $[+11.23,+16.82]$~pp). At 30\% coverage, the corresponding gains were still 4.67 and 5.47~pp, with both confidence intervals remaining above zero. Synthetic completion therefore provided a substantial advantage when most identities would otherwise be absent from cross-domain training.

The effect became progressively smaller at higher genuine coverage. At 50\% and 70\%, the mixed conditions exceeded the A controls by 1.65 and 1.48~pp, respectively, but both confidence intervals included zero. Relative to the B controls, the corresponding effects were 2.27 and 2.29~pp, with lower confidence bounds just above zero. At 90\% coverage, the mixed, A, and B conditions converged and the paired effects were close to zero. The results therefore support a diminishing-return interpretation: synthetic completion is most valuable when genuine historical observations are severely limited and contributes progressively less as the missing portion of the old domain becomes small.

The B controls help distinguish this completion effect from additional optimization. At every evaluated coverage level, the B--A point estimate was small and negative, and every confidence interval included zero. Increasing the number of optimization steps and repeatedly sampling the same reduced genuine identity set therefore did not reproduce the gain obtained by retaining the full training identity set through synthetic completion. The result should not be interpreted as evidence that repeated exposure is detrimental; rather, there is no indication that additional exposure alone explains the completion advantage.

The acceptable-loss analysis provides a complementary practical perspective. At 50\% and 60\% genuine coverage, the point estimates were already close to the complete-genuine-data reference, but the lower confidence bounds extended beyond the prespecified two-percentage-point margin. At 70\% coverage, the paired difference relative to \texttt{R100\_S0} was $+0.03$~pp with a 95\% CI of $[-1.78,+1.88]$~pp. Thus, 70\% was the lowest evaluated genuine coverage level whose hierarchical confidence interval remained within the prespecified acceptable-loss margin. This value should be interpreted as an empirical result for the present dataset and protocol rather than as a universal requirement for cultural heritage retrieval.

\subsection{Identity coverage and augmentation effects}
\label{sec:discussion_mechanisms}

The control experiments support broader training identity coverage as an important mechanism behind synthetic completion, particularly under severe scarcity. Mixed conditions retain all training identities and provide an old--new pair for each of them, whereas the A and B controls remove identities without genuine historical observations entirely. At 10\% coverage, for example, the mixed model is trained across the full identity set, while the real-only controls retain only approximately one tenth of those identities. The large mixed--A and mixed--B effects at this coverage level are therefore consistent with the value of preserving a broad set of cross-domain training identities rather than repeatedly optimizing on a much smaller genuine subset.

This interpretation is also supported by the rapid reduction of the completion effect as genuine coverage increases. When most identities already contain genuine old-domain observations, synthetic completion adds progressively fewer otherwise missing identities to the cross-domain training set. By 90\% genuine coverage, little difference remains between mixed and real-only training.

The results do not, however, support attributing all benefits of synthetic data exclusively to identity completion. In the \textsc{Real+Synth} condition, all identities already had genuine old-domain observations, yet adding OLDIFY1 increased mean R@1 from 92.15\% to 94.05\%. The paired effect was $+1.90$~pp, with a 95\% CI of $[-0.10,+4.09]$~pp. Because the interval narrowly
included zero, this result does not establish a clear augmentation benefit, but it provides a signal that synthetic observations may also contribute useful appearance variability independently of missing-identity completion.

The relative importance of these mechanisms appears to depend on the degree of scarcity. At 10\% and 30\% genuine coverage, the completion effects of approximately 5--14~pp are much larger than the full-coverage \textsc{Real+Synth} effect, indicating that preservation of broader training identity coverage is the dominant explanation in this regime. At 50\%--70\% coverage, the completion effects are of roughly the same order as the \textsc{Real+Synth} effect. Both identity completion and more general augmentation may therefore contribute when genuine historical coverage is already moderate.

\subsection{Dataset composition and retrieval implications}
\label{sec:discussion_dataset_composition}

Absolute retrieval performance and the magnitude of the completion effect varied across dataset partitions, indicating that the value of historical observations depends not only on their number but also on the identities represented in the training and test sets. The decreasing completion effect with increasing genuine coverage was, however, reproduced across all three pairwise disjoint test partitions. At 10\% and 30\% genuine coverage, the mixed condition outperformed its exposure-matched A control in every partition, whereas the differences became much smaller at 50\% and 70\% and effectively disappeared at 90\%.

This variability suggests that genuine historical observations have nonuniform informational value. Records involving large viewpoint changes, unusual degradation, strong cropping, uncommon reproduction processes, or visually ambiguous objects may provide more useful cross-domain supervision than records differing mainly in low-level appearance. Consequently, subsets containing the same number of historical identities need not be equally informative.

From a collection-management perspective, the results suggest a hybrid strategy. Genuine historical observations remain the most valuable source of cross-domain supervision, but synthetic completion can reduce the performance cost of incomplete archival coverage. Rather than requiring genuine historical material for every training identity, an informative genuine subset could potentially be combined with synthetic counterparts for identities lacking archival records. The present study does not determine how such a genuine subset should be selected, but retrieval difficulty, domain discrepancy, visual diversity, or expected information gain are plausible criteria for future investigation.

The high R@5 and R@10 values further distinguish expert-assisted from fully automatic retrieval. The strongest effects of historical-data scarcity occur at rank~1, whereas the correct identity generally remains within a short candidate list at greater retrieval depths. Synthetic training may therefore be particularly useful in workflows where an automated system narrows a large collection to a small set of candidates that can subsequently be inspected by a curator or other domain expert.

More broadly, synthetic data can be viewed as a mechanism for completing an incompletely observed domain rather than solely as a means of enlarging an already represented image distribution. This perspective may apply to other instance-level cross-domain retrieval tasks in which one observation domain is widely available but its counterpart exists only for a subset of identities.

\subsection{Limitations}
\label{sec:limitations}

Several limitations constrain the generalization and causal interpretation of these findings. First, the experiments use one dataset assembled from two cultural heritage sources. Although the three test sets are pairwise identity-disjoint and together contain 483 distinct test identities, all observations originate from the same assembled collection, and training or validation identities may occur in more than one partition. The observed relationship between genuine historical coverage and synthetic completion should therefore be validated on independent collections with different object types, acquisition processes, and degrees of domain shift.

Second, the study evaluates one retrieval backbone, one metric-learning objective, and one degradation-based synthetic aging pipeline. Different architectures, loss functions, sampling strategies, or synthetic-generation methods may respond differently to the same scarcity regime. In particular, OLDIFY mainly modifies image appearance while preserving much of the viewpoint, framing, composition, and geometry of its source. Synthetic methods capable of introducing realistic structural or acquisition changes may reduce the remaining gap between synthetic and genuine historical observations.

Third, the A and B controls do not isolate cross-domain pairing from overall identity diversity completely. Identities without genuine old-domain observations are removed entirely from these controls, including their contemporary images. The mixed variants therefore differ from A and B not only in the availability of old-domain counterparts, but also in the number of unique identities retained in training. The results are consequently consistent with broader cross-domain training identity coverage being a primary contributor, but the present design cannot determine how much of the gain arises specifically from restored old--new pairing and how much arises from retaining a larger and more diverse identity set.

Fourth, the A variants match expected per-identity exposure only up to rounding and do not reproduce complete optimization trajectories. Early stopping and stochastic sampling can result in different total numbers of updates across trained models. The B variants provide a complementary full-budget control, but necessarily increase repeated sampling of the reduced genuine subsets. Together, the controls show that additional exposure alone does not reproduce the large low-coverage completion gains, but they cannot remove every possible interaction between identity-set size and optimization.

Fifth, scarcity is simulated by selecting nested genuine identity subsets rather than by reproducing the historical missingness process of a particular museum or archive. Although the subsets are constructed using category-stratified ordering, real collections may lack historical documentation systematically rather than randomly. Objects with certain ages, materials, acquisition histories, or documentation practices may be more likely to have archival images. The evaluated coverage percentages should therefore not be interpreted
as universal thresholds applicable to all collections.

Finally, confidence intervals are estimated using paired bootstrap resampling at the test-identity level together with resampling of training seeds. Identity-level resampling preserves the dependence among queries belonging to the same object in the small number of cases with multiple images, while the three evaluated partitions retain equal weight. The resulting confidence intervals quantify uncertainty over the evaluated test identities and training seeds; they should not be interpreted as population-level uncertainty across independent cultural heritage datasets. Similarly, the prespecified two-percentage-point acceptable-loss criterion provides a practical descriptive reference rather than a formal non-inferiority test.

\section{Conclusions}
\label{sec:conclusions}

Synthetic aging cannot fully replace genuine historical training data in cross-domain cultural heritage retrieval. Complete replacement reduced bidirectional mean R@1 from 92.15\% to 87.94\%, corresponding to a paired effect of $-4.21$ percentage points. Increasing the synthetic sampling pool from one to three independently generated variants did not improve retrieval,
indicating that additional stochastic realizations of the same degradation pipeline do not reproduce the variability contained in genuine historical observations.

Synthetic completion was substantially more effective when genuine historical observations were scarce. At 10\% genuine old-domain identity coverage, retaining the full training identity set through synthetic completion improved mean R@1 by 12.30~pp relative to the exposure-matched real-only control and by 13.99~pp relative to the real-only control using the full optimization budget. At 30\% coverage, the corresponding gains were 4.67 and 5.47~pp. The completion advantage decreased as genuine historical coverage increased and was effectively absent at 90\%.

Increasing repeated exposure to the same reduced genuine identity sets did not reproduce these gains. The results are therefore consistent with broader cross-domain training identity coverage being a primary contributor to the value of synthetic completion, particularly under severe scarcity. The present controls do not, however, separate this effect completely from the benefit of retaining a larger and more diverse set of training identities. 

Using a prespecified acceptable-loss margin of two percentage points, 70\% genuine old-domain identity coverage was the lowest evaluated level whose 95\% bootstrap confidence interval remained within the margin relative to complete genuine coverage. This value should be interpreted as specific to the present dataset and experimental protocol rather than as a universal requirement for cultural heritage retrieval.

Synthetic data may also provide a weaker augmentation effect when genuine historical coverage is already complete. Adding OLDIFY1 observations to the complete genuine training set increased mean R@1 from 92.15\% to 94.05\%, although the paired 95\% confidence interval ($[-0.10,+4.09]$~pp) narrowly included zero. Genuine and synthetic observations should therefore be regarded as
complementary rather than interchangeable: genuine images provide authentic historical variability, while synthetic observations can extend cross-domain supervision where archival records are missing and may additionally contribute useful appearance variability.

For cultural heritage applications, the practical implication is that synthetic data are most valuable when used to fill specific informational gaps rather than simply to increase image count. Training-set construction should therefore consider which identities and domains are missing, not only the total number of available images. This perspective can support artwork matching, collection integration, catalog enrichment, provenance research, and other retrieval tasks involving heterogeneous historical documentation.

\section*{Declaration of competing interest}
The authors declare that they have no known competing financial interests or personal relationships that could have appeared to influence the work reported in this paper.

\section*{Data availability}
The source images used in this study originate from the publicly accessible Louvre Collections and Lost Art Database. Access to, reuse of, and redistribution of these images remain subject to the terms and copyright restrictions of the respective source institutions. The dataset partitions, experimental configurations, aggregated results, and source code used to generate the synthetic images and evaluate the retrieval models are available from the corresponding author upon reasonable request.

\section*{Funding}
This work was supported by the Polish Agency for Enterprise Development (PARP) under Grant No. FENG.01.01-IP.02-3717/23, co-funded by the European Union through the European Funds for a Modern Economy 2021–2027 (FENG), SMART Path.

\section*{Declaration of generative AI and AI-assisted technologies in the manuscript preparation process}

During the preparation of this work, the authors used OpenAI ChatGPT to support data analysis, language editing, restructuring, and improvement of the clarity and readability of the manuscript. After using this tool, the authors reviewed, verified, and edited the content as needed and take full responsibility for the content of the published
article.
\bibliographystyle{elsarticle-num}
\bibliography{references}

\noindent \textbf{\large Supplementary Material}
\setcounter{section}{0}
\setcounter{table}{0}
\renewcommand{\thesection}{S\arabic{section}}
\renewcommand{\thetable}{S\arabic{table}}
\setcounter{figure}{0}
\renewcommand{\thefigure}{S\arabic{figure}}
\renewcommand{\theHfigure}{S\arabic{figure}}

\section{Synthetic aging procedure}
\label{sup:synthetic_aging}

The OLDIFY images were generated offline from contemporary images in the training subset. Images whose larger spatial dimension exceeded 1500 pixels were proportionally downscaled to a maximum dimension of 1500 pixels; smaller images retained their original resolution.
 
For each generated image, all transformation parameters were sampled independently from the ranges in Table~\ref{tab:oldify_parameters}. Continuous parameters followed \(\mathcal{U}(a,b)\), whereas integer parameters followed a discrete uniform distribution including both endpoints. All processing stages were enabled, although an effect could become negligible when its sampled intensity was zero or close to zero.
 
OLDIFY1, OLDIFY2, and OLDIFY3 were independent realizations of the same stochastic procedure. They used identical transformation definitions, ranges, and ordering, differing only in the sampled parameter values. 

\begin{table}[htbp]
\centering
\scriptsize
\caption{Parameter ranges used for synthetic aging. Parameters were sampled
independently for each generated image.}
\label{tab:oldify_parameters}
\begin{tabular}{lll}
\hline
Parameter & Sampling range & Interpretation \\
\hline
\texttt{frame\_thickness}
    & $\mathcal{U}_{\mathrm{int}}(4,15)$ px
    & Replicated border \\

\texttt{gaussian\_mean}
    & $\mathcal{U}(0.00,0.05)$
    & Gaussian-noise mean \\

\texttt{gaussian\_var}
    & $\mathcal{U}(0.001,0.010)$
    & Gaussian-noise variance \\

\texttt{salt\_pepper\_amount}
    & $\mathcal{U}(0.001,0.005)$
    & Impulse-noise fraction per polarity \\

\texttt{vignette\_strength}
    & $\mathcal{U}(0.55,0.95)$
    & Vignetting strength \\

\texttt{scratch\_intensity}
    & $\mathcal{U}_{\mathrm{int}}(0,4)$
    & Scratch and dust intensity \\

\texttt{blur\_sigma}
    & $\mathcal{U}(1.0,3.0)$
    & Gaussian-blur standard deviation \\

\texttt{gamma}
    & $\mathcal{U}(0.85,1.20)$
    & Gamma correction \\

\texttt{perspective\_top\_shift}
    & $\mathcal{U}(0.00,0.06)$
    & Relative top-edge inset \\

\texttt{perspective\_bottom\_shift}
    & $\mathcal{U}(0.00,0.06)$
    & Relative bottom-edge inset \\

\texttt{perspective\_left\_shift}
    & $\mathcal{U}(0.00,0.15)$
    & Relative left-edge inset \\

\texttt{perspective\_right\_shift}
    & $\mathcal{U}(0.00,0.15)$
    & Relative right-edge inset \\

\texttt{paper\_strength}
    & $\mathcal{U}(0.08,0.25)$
    & Paper-texture modulation \\

\texttt{leak\_intensity}
    & $\mathcal{U}(0.00,0.25)$
    & Light-leak intensity \\

\texttt{leak\_angle\_deg}
    & $\mathcal{U}(0,360)$ deg
    & Light-leak origin \\

\texttt{sepia\_strength}
    & $\mathcal{U}(0.90,1.00)$
    & Sepia-to-grayscale interpolation \\
\hline
\end{tabular}
\end{table}

\subsection{Processing sequence}

The transformations were applied in the following fixed order:

\begin{enumerate}
    \item \textbf{Replicated border.} A border of 4--15 pixels was added by replicating the outermost image pixels.

    \item \textbf{Gaussian and impulse noise.} Additive Gaussian noise was applied in the normalized intensity range \([0,1]\), after which approximately \(p_{\mathrm{sp}}HW\) pixels were
    independently replaced with white values and the same number with black values. Intensities were clipped to the valid range.

    \item \textbf{Vignetting.} Each channel was multiplied by \(V(x,y)=1-s_{\mathrm{v}}r(x,y)^2\), where \(r(x,y)\) is the normalized distance from the image center. The mask was clipped to \([0,1]\). 

    \item \textbf{Scratches and dust.} For positive intensity values, randomly positioned lines of 20--80 pixels in length and one or two pixels in thickness were generated and blurred.
    Sparse dust artifacts were added with density proportional to the sampled intensity. This stage was inactive when the intensity equaled zero.

    \item \textbf{Blur and gamma correction.} Gaussian blur was followed by the transformation \(I'=I^{1/\gamma}\) in normalized intensity space.

    \item \textbf{Perspective transformation.} The four image corners were moved toward the interior according to the independently sampled top, bottom, left, and right offsets. The resulting
    quadrilateral was projected onto a rectangular image. Only its internal region was retained, so the output could be smaller than the input.

    \item \textbf{Paper texture.} Uniform random noise was smoothed with a Gaussian filter of \(\sigma=3\), normalized to \([0.85,1.15]\), and applied as \(I'=I\,T^{s_{\mathrm{paper}}}\).

    \item \textbf{Light leak.} A warm artifact was generated from an image corner determined by the sampled angle. Its intensity decreased quadratically with distance from
    the origin. Relative BGR channel contributions were \(0.40\), \(0.65\), and \(1.00\).

    \item \textbf{Sepia-to-grayscale conversion.}    A standard sepia transform was applied using

    \[
    \begin{bmatrix}
    R_{\mathrm{s}}\\
    G_{\mathrm{s}}\\
    B_{\mathrm{s}}
    \end{bmatrix}
    =
    \begin{bmatrix}
    0.393 & 0.769 & 0.189\\
    0.349 & 0.686 & 0.168\\
    0.272 & 0.534 & 0.131
    \end{bmatrix}
    \begin{bmatrix}
    R\\
    G\\
    B
    \end{bmatrix}.
    \]

    The result was interpolated with its grayscale version using \(\alpha=2s_{\mathrm{sepia}}-1\) and \(I_{\mathrm{out}}=(1-\alpha)I_{\mathrm{sepia}} + \alpha I_{\mathrm{gray}}\). Because
    \(s_{\mathrm{sepia}}\in[0.90,1.00]\), the generated images ranged from strongly desaturated sepia to fully grayscale.
\end{enumerate}

OLDIFY1 was used in the controlled-scarcity and synthetic-completion experiments. The union of OLDIFY1, OLDIFY2, and OLDIFY3 was used only in the \textsc{Synth-3} complete-replacement condition. \section{Partition-level retrieval results}
\label{sup:partition_results}

Partition-level results were examined to assess whether the relationship between genuine old-domain coverage and synthetic completion was reproduced across different test-set compositions. Each partition contains 161 test identities, and the three test identity sets are pairwise disjoint. Within each partition, the reported values are averaged over the three training seeds.

\begin{figure}[htbp]
    \centering
    \includegraphics[width=0.5\textwidth]
    {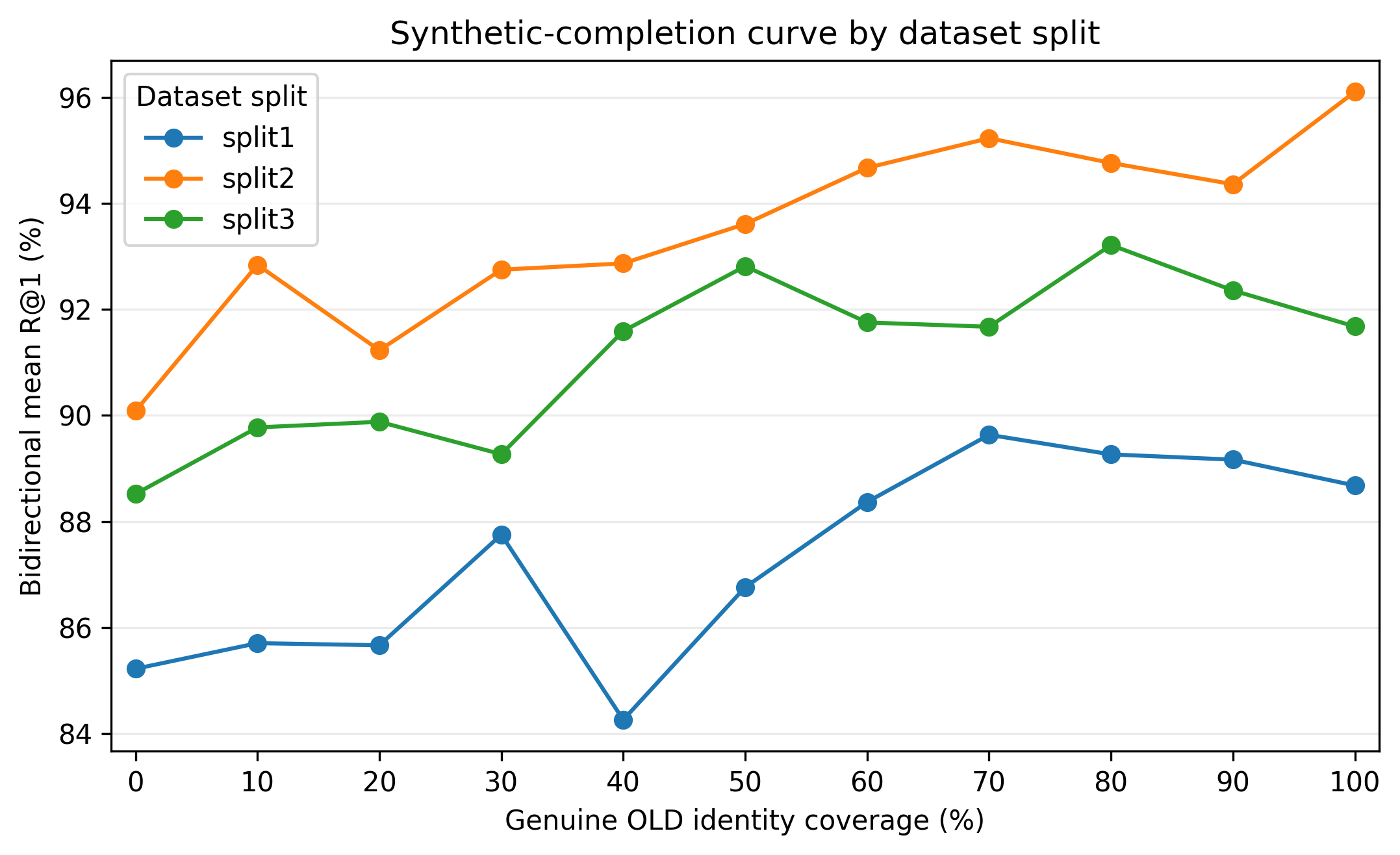}
    \caption{Bidirectional mean R@1 for the real--synthetic completion series, reported separately for the three dataset partitions and averaged over three training seeds. Each curve spans genuine old-domain identity coverage from 0\% to 100\%; identities without genuine old-domain observations are completed using OLDIFY1 images.}
    \label{fig:sup_split_completion_curves}
\end{figure}

As shown in Figure~\ref{fig:sup_split_completion_curves}, absolute retrieval performance differed across partitions, indicating differences in test-set difficulty. Nevertheless, the general relationship between genuine historical coverage and retrieval performance was similar: synthetic-completion performance improved substantially from the fully synthetic endpoint and tended to flatten at higher genuine-coverage levels. The curves were not strictly monotonic within individual partitions.

The benefit of synthetic completion relative to the exposure-matched real-only A controls was particularly consistent under severe scarcity. At 10\% genuine coverage, the mixed--A differences were $+10.09$, $+14.63$, and $+12.19$ percentage points (pp) for Splits~1, 2, and 3, respectively. At 30\%, the corresponding differences were $+6.29$, $+4.36$, and $+3.35$~pp.
Thus, the large completion benefit observed at low genuine coverage was reproduced in all three pairwise disjoint test partitions.

At intermediate coverage, the magnitude of the effect became smaller but remained positive in each partition. At 50\% genuine coverage, the mixed--A differences were $+0.87$, $+1.30$, and $+2.79$~pp, while at 70\% they were $+0.76$, $+2.06$, and $+1.63$~pp for Splits~1--3, respectively. By 90\% genuine coverage, the differences were small and their direction was no
longer consistent across partitions.

These results indicate that dataset composition affects both absolute retrieval difficulty and effect magnitude, but the main scarcity-related pattern is stable: synthetic completion provides its largest and most consistent benefit when genuine historical observations cover only a small fraction of the training identities, and this benefit diminishes as genuine coverage approaches completeness.

\section{Detailed and secondary retrieval results}
\label{sup:detailed_results}

This section provides numerical results complementary to the primary analyses reported in the main text. It includes the complete genuine-coverage series and real-only controls, the prespecified acceptable-loss comparison with complete genuine old-domain coverage, and retrieval performance at larger rank depths.

\subsection{Complete scarcity-series results}
\label{sup:complete_scarcity_results}

Table~\ref{tab:supp_scarcity_results} reports bidirectional mean R@1 for the complete real--synthetic completion series and the real-only A and B controls. Values are averages over three dataset partitions and three training seeds. The mixed series retains all 808 training identities and replaces missing genuine old-domain observations with OLDIFY1 images. The A and B controls retain only identities with genuine old-domain observations.

\begin{table}[htbp]
\centering
\scriptsize
\caption{Bidirectional mean R@1 for the complete controlled-scarcity series and real-only controls, averaged over three dataset partitions and three training seeds. Mixed conditions use OLDIFY1 observations for identities without genuine old-domain images. A controls use matched expected per-identity exposure, whereas B controls use the full per-epoch optimization budget.}
\label{tab:supp_scarcity_results}
\begin{tabular}{lcc}
\hline
Condition &
Genuine OLD coverage &
Mean R@1 \\
\hline
\texttt{R0\_S100}   & 0\%   & 87.94\% \\
\texttt{R10\_S90}   & 10\%  & 89.44\% \\
\texttt{R10\_A}     & 10\%  & 77.14\% \\
\texttt{R10\_B}     & 10\%  & 75.45\% \\
\hline
\texttt{R20\_S80}   & 20\%  & 88.92\% \\
\hline
\texttt{R30\_S70}   & 30\%  & 89.92\% \\
\texttt{R30\_A}     & 30\%  & 85.26\% \\
\texttt{R30\_B}     & 30\%  & 84.45\% \\
\hline
\texttt{R40\_S60}   & 40\%  & 89.57\% \\
\hline
\texttt{R50\_S50}   & 50\%  & 91.06\% \\
\texttt{R50\_A}     & 50\%  & 89.41\% \\
\texttt{R50\_B}     & 50\%  & 88.79\% \\
\hline
\texttt{R60\_S40}   & 60\%  & 91.59\% \\
\hline
\texttt{R70\_S30}   & 70\%  & 92.18\% \\
\texttt{R70\_A}     & 70\%  & 90.69\% \\
\texttt{R70\_B}     & 70\%  & 89.89\% \\
\hline
\texttt{R80\_S20}   & 80\%  & 92.41\% \\
\hline
\texttt{R90\_S10}   & 90\%  & 91.96\% \\
\texttt{R90\_A}     & 90\%  & 92.18\% \\
\texttt{R90\_B}     & 90\%  & 92.04\% \\
\hline
\texttt{R100\_S0}   & 100\% & 92.15\% \\
\hline
\end{tabular}
\end{table}

The mixed completion series shows an overall increase in retrieval performance with increasing genuine old-domain coverage, but the relationship is not strictly monotonic. The largest improvements relative to the real-only controls occur at low genuine coverage. Differences become progressively smaller at intermediate coverage and are negligible at 90\%, consistent with the paired analyses in the main text. Small discrepancies between differences calculated from the rounded values in Table~\ref{tab:supp_scarcity_results} and the reported paired effects arise because statistical calculations use unrounded query-level results.

\subsection{Performance relative to complete genuine coverage}
\label{sup:acceptable_loss}

\begin{figure}[htbp]
    \centering
    \includegraphics[width=0.5\textwidth]
    {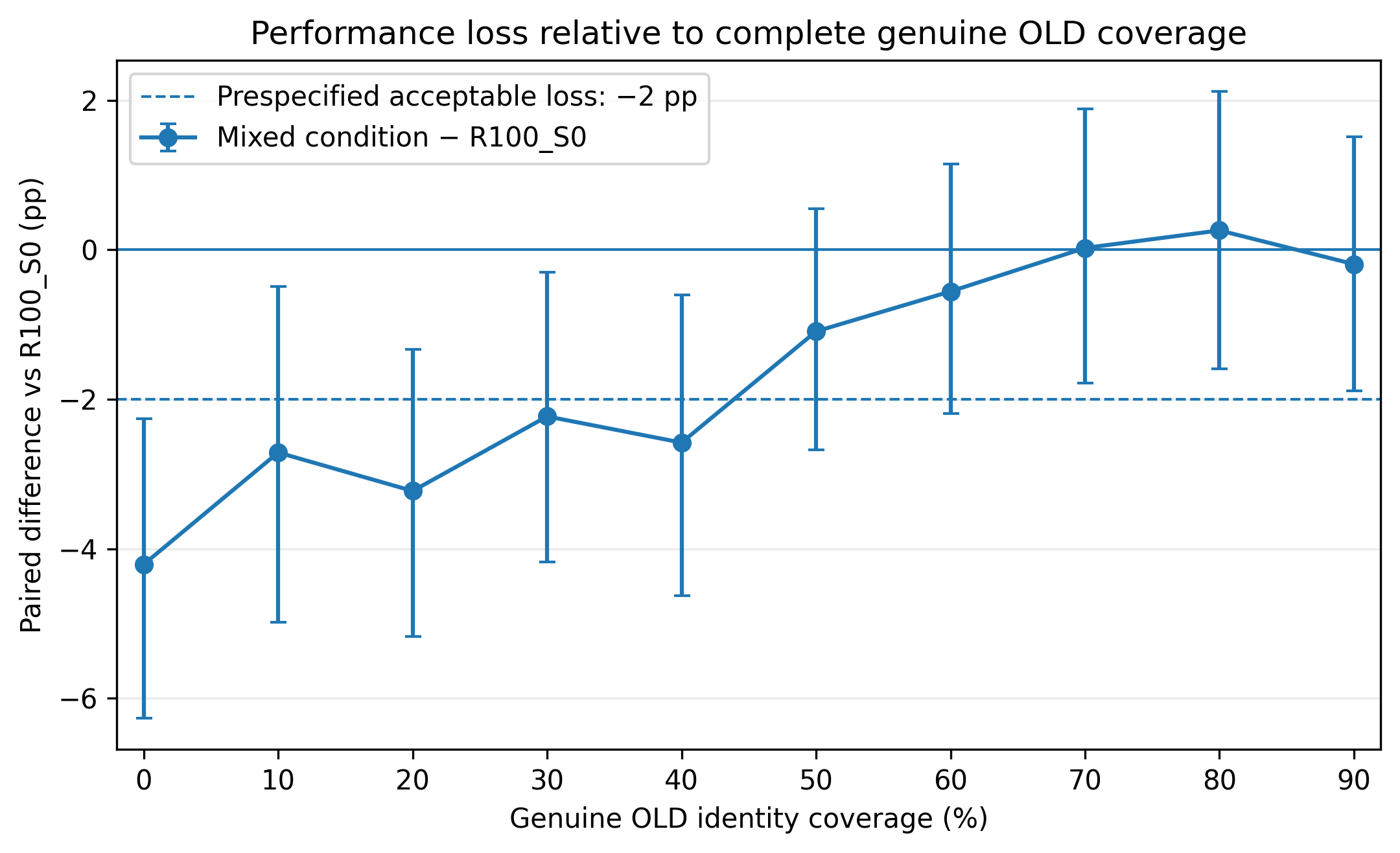}
    \caption{Paired difference in bidirectional mean R@1 between
    each real--synthetic completion condition and the complete
    genuine-data reference \texttt{R100\_S0}. Error bars represent 95\% confidence intervals obtained using the paired identity-level bootstrap described in Section 3.6.}
    \label{fig:sup_completion_vs_r100}
\end{figure}

The mixed completion conditions were compared directly with the complete genuine-data reference \texttt{R100\_S0}. Figure~\ref{fig:sup_completion_vs_r100} shows the paired difference in
bidirectional mean R@1 and the prespecified acceptable-loss boundary of $-2$ percentage points.

For completeness, the numerical paired effects are reported in Table~\ref{tab:supp_acceptable_loss}. A condition meets the descriptive acceptable-loss criterion when the lower bound of its 95\% confidence interval is greater than $-2$~pp.

\begin{table}[htbp]
\centering
\scriptsize
\caption{Paired differences in bidirectional mean R@1 between the real--synthetic completion series and the complete genuine-data reference \texttt{R100\_S0}. Confidence intervals were estimated using the paired identity-level bootstrap with seed resampling described in Section 3.6.}
\label{tab:supp_acceptable_loss}
\begin{tabular}{lccc}
\hline
Genuine OLD coverage &
Difference vs.\ \texttt{R100\_S0} &
95\% CI &
2-pp criterion \\
\hline
0\%  & $-4.21$ pp & $[-6.19,-2.23]$ pp & No \\
10\% & $-2.71$ pp & $[-4.98,-0.49]$ pp & No \\
20\% & $-3.23$ pp & $[-5.18,-1.33]$ pp & No \\
30\% & $-2.23$ pp & $[-4.18,-0.30]$ pp & No \\
40\% & $-2.58$ pp & $[-4.63,-0.60]$ pp & No \\
50\% & $-1.09$ pp & $[-2.68,+0.55]$ pp & No \\
60\% & $-0.56$ pp & $[-2.19,+1.15]$ pp & No \\
70\% & $+0.03$ pp & $[-1.78,+1.88]$ pp & Yes \\
80\% & $+0.26$ pp & $[-1.59,+2.12]$ pp & Yes \\
90\% & $-0.19$ pp & $[-1.88,+1.52]$ pp & Yes \\
\hline
\end{tabular}
\end{table}

The point estimates approached the complete-genuine-data reference before the uncertainty intervals satisfied the prespecified margin. At 50\% and 60\% genuine coverage, the mean differences were only $-1.09$ and $-0.56$~pp, respectively, but their lower confidence bounds extended below $-2$~pp. At 70\%, the paired difference was $+0.03$~pp with a 95\% CI of $[-1.78,+1.88]$~pp. Thus, 70\% was the lowest evaluated genuine-coverage level satisfying the descriptive 2-pp acceptable-loss criterion. The 80\% and 90\% conditions also satisfied the criterion. As stated in the main text, this analysis is not interpreted as a formal non-inferiority test.

\subsection{Rank-dependent retrieval performance}
\label{sup:rank_depth}

\begin{figure}[htbp]
    \centering
    \includegraphics[width=0.5\textwidth]
    {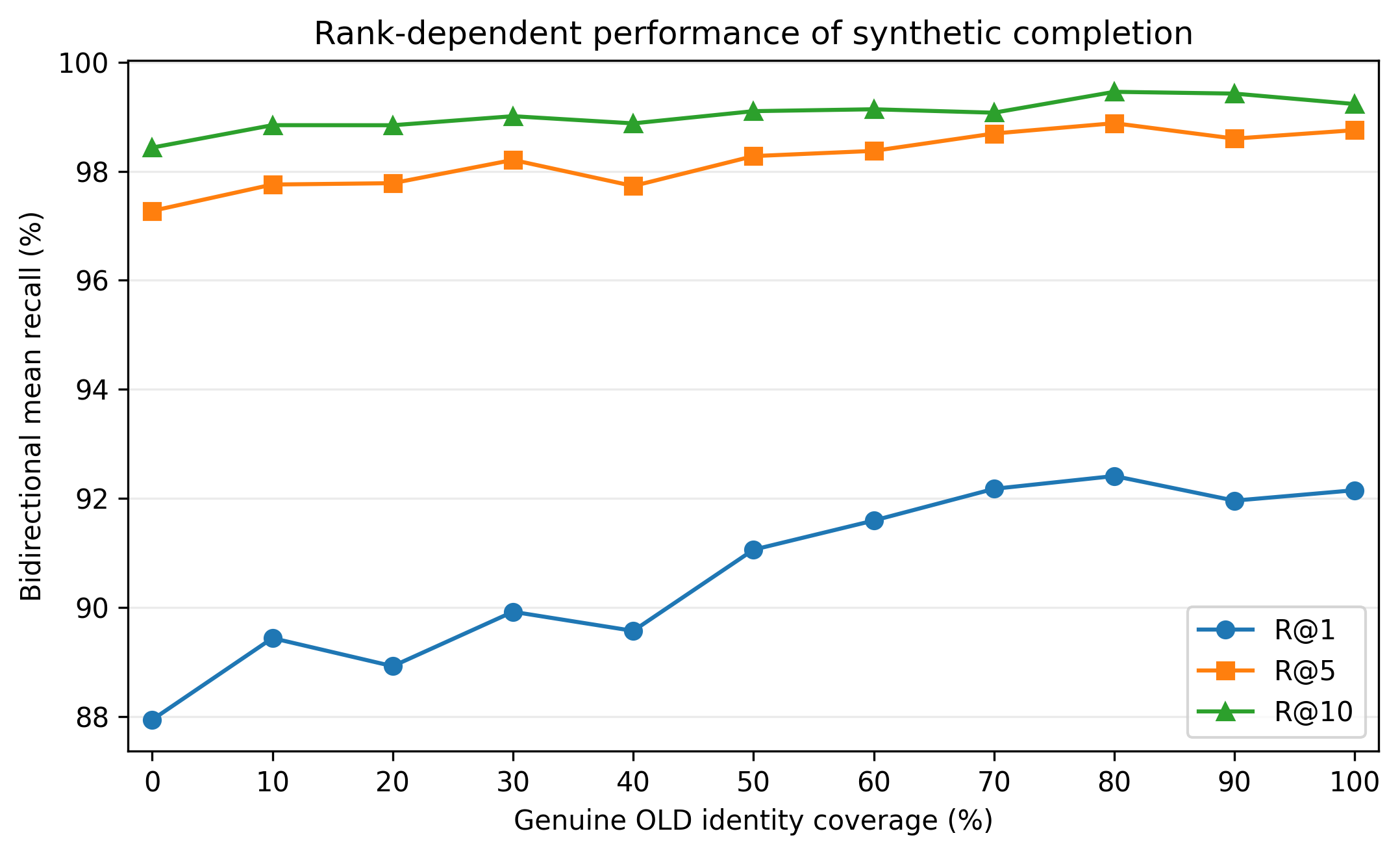}
    \caption{Rank-dependent retrieval performance of the real--synthetic completion series. Curves show bidirectional mean R@1, R@5, and R@10 as a function of genuine old-domain identity coverage, averaged over three dataset partitions and three
    training seeds.}
    \label{fig:sup_rank_depth}
\end{figure}

Figure~\ref{fig:sup_rank_depth} compares bidirectional mean recall at ranks 1, 5, and 10 across the complete real--synthetic completion
series.

The effect of historical-data scarcity was substantially larger at rank~1 than at greater retrieval depths. Even complete synthetic replacement achieved 98.43\% bidirectional mean R@10, while the complete genuine-data reference reached 99.23\%. Across the mixed series, R@10 remained close to 99\%, whereas the corresponding R@1 values varied more strongly with genuine old-domain coverage.

These results indicate that synthetic aging often preserves enough identity information to place the correct object within a small set of highly ranked candidates, even when it does not reproduce the genuine historical domain sufficiently well to rank the correct identity first. The distinction supports the potential use of synthetic completion in expert-assisted retrieval workflows, where the automated system provides a short candidate list for subsequent manual verification.

\end{document}